\documentclass[11pt,a4paper]{article}
\usepackage{authblk}
\usepackage[hyperref]{acl2020}
\usepackage{times}
\usepackage{latexsym}

\usepackage{microtype}

\aclfinalcopy 

\usepackage{url}
\usepackage{graphicx}
\usepackage{diagbox}
\usepackage{multirow}
\usepackage{multicol}
\usepackage{amsmath}
\usepackage{subcaption}

\makeatletter
\def\blfootnote{\xdef\@thefnmark{}\@footnotetext}
\makeatother

\title{Speaker-change Aware CRF for Dialogue Act Classification}

\author[1,2]{Guokan Shang}
\author[1]{Antoine J.-P. Tixier}
\author[1,3]{\\Michalis Vazirgiannis}
\author[2]{Jean-Pierre Lorr\'e}
\affil[1]{\'Ecole Polytechnique, $^\mathrm{2}$Linagora, $^\mathrm{3}$AUEB}

\date{}

\begin{document}
\maketitle
\begin{abstract}
Recent work in Dialogue Act (DA) classification approaches the task as a sequence labeling problem, using neural network models coupled with a Conditional Random Field (CRF) as the last layer.
CRF models the conditional probability of the target DA label sequence given the input utterance sequence.
However, the task involves another important input sequence, that of speakers, which is ignored by previous work.
To address this limitation, this paper proposes a simple modification of the CRF layer that takes speaker-change into account.
Experiments on the SwDA corpus show that our modified CRF layer outperforms the original one, with very wide margins for some DA labels.
Further, visualizations demonstrate that our CRF layer can learn meaningful, sophisticated transition patterns between DA label pairs conditioned on speaker-change in an end-to-end way.
Code is publicly available\footnote{\url{ https://bitbucket.org/guokan_shang/da-classification}}.
\end{abstract}

\blfootnote{GS handled the data, implemented the model, ran the experiments, and generated the plots.
GS and AJPT equally participated in the design of the study and the writing of the paper.}

\section{Introduction}
A conversation can be seen as a sequence of utterances.
The task of Dialogue Act (DA) classification aims at assigning to each utterance a DA label to represent its communicative intention.
Dialogue acts originate from the notion of \textit{illocutionary force} (speaker's intention in delivering an utterance) introduced back in the theory of Speech Act \citep{austin1962things,searle1969speech}.
DAs are assigned based on a combination of syntactic, semantic, and pragmatic criteria \citep{stolcke-etal-2000-dialogue}.
As shown in Table \ref{table:conversation_sample}, some examples of DAs include stating, questioning, answering, etc.
The full set of DA labels is predefined.
A number of annotation schemes have been developed, varying from domain-specific, such as VERBMOBIL \citep{alexanderssony1997dialogue}, to domain-independent, such as DAMSL \citep{allen1997draft, core1997coding} and DiAML\footnote{accepted to be included in the ISO 24617-2 standard.  \url{https://www.iso.org/standard/76443.html}} \citep{bunt-etal-2010-towards,bunt-etal-2012-iso}.

Automatically detecting DA labels is an essential step towards describing the discourse structure of conversation \citep{jurafsky1997switchboard}.
DAs are very useful annotations to a large variety of spoken language understanding tasks, such as utterance clustering \citep{shang2019energy}, real-time information retrieval \cite{meladianos2017real}, conversational agents \citep{higashinaka-etal-2014-towards,ahmadvand2019contextual}, and summarization \cite{shang2018unsupervised}.

\begin{table}[t]
\small
\centering
\setlength{\tabcolsep}{3pt} 
\renewcommand{\arraystretch}{1.1} 
\scalebox{0.9}{
\begin{tabular}{ccl|c}
\hline
	 Change & Speaker & Utterance & DA\\
\hline
     - & B & Of course I use, & sd  \\
     True & A & $<$laughter$>$. & x \\
     True & B & credit cards. & + \\
     False & B & I have a couple of credit cards & sd \\
     True & A & Yeah. & b \\
     True & B & and, uh, use them. & + \\
     True & A & Uh-huh, & b\\
     False & A & do you use them a lot? & qy\\
     True & B & Oh, we try not to. & ng\\
\hline
\end{tabular}
}
\caption{Fragment from SwDA conversation sw3332. Statement-non-opinion (sd), Non-verbal (x), Interruption (+), Acknowledge/Backchannel (b), Yes-No-Question (qy), Negative non-no answers (ng).}
\label{table:conversation_sample}
\end{table}

It is difficult to predict the DA of a single utterance without having access to the other utterances in the context. For instance, for an utterance such as ``{\small\texttt{Yeah}}'', it is hard to tell whether the associated DA should be `Agreement', `Yes answer' or `Backchannel'. 
Plus, different labels have different transition probabilities to other labels.
E.g., an initial greeting DA is very likely to be followed by another greeting DA.
Likewise, a question DA is more likely to be followed by an answer DA.
To summarize, it is necessary for a DA classification model to capture dependencies both at the utterance level and at the label level.
Recent works \citep{li-wu-2016-multi,tran-etal-2017-hierarchical,liu-etal-2017-using-context,kumar2018dialogue,chen2018dialogue,raheja-tetreault-2019-dialogue,li-etal-2019-dual} treat DA classification as a sequence labeling problem.
The BiLSTM-CRF model \citep{huang2015bidirectional,lample-etal-2016-neural}, originally introduced for the tasks of POS tagging, chunking and named entity recognition, is the most widely used architecture.
In it, a bidirectional recurrent neural network with LSTM cells is first applied to capture the dependencies among consecutive utterances, and then, a Conditional Random Field (CRF) layer is used to capture the dependencies among consecutive DA labels.

CRF is a discriminative probabilistic graphical framework \citep{koller2009probabilistic,sutton2012introduction} used to label sequences \citep{lafferty2001conditional}.
It models the conditional probability of a target label sequence given an input sequence.
General CRF can essentially model any kind of graphical structure to capture arbitrary dependencies among output variables.
For NLP sequence labeling tasks, linear chain CRF is the most common variant.
The labels are arranged in a linear chain, i.e., only neighboring labels are dependent (first-order Markov assumption).
The BiLSTM-CRF architecture employs a linear chain CRF.
Hence, for brevity, in the rest of this paper, the term CRF is short for linear chain CRF.

Recently, neural versions of the CRF have been developed mainly for NLP sequence labeling tasks \citep{collobert2011natural,huang2015bidirectional,lample-etal-2016-neural}. 
While traditional CRF requires defining a potentially large set of handcrafted feature functions (each weighted with a parameter to be trained), neural CRF has only two parameterized feature functions (emission and transition) that are trained with the other parameters of the network in an end-to-end fashion.

\section{Motivation}
Most sequence labeling tasks in NLP, such as POS tagging, chunking, and named entity recognition, involve only two sequences: input and target.
In DA classification however, we have access to an additional input sequence, that of speaker-identifiers.
This extra input could, in principle, greatly improve DA prediction.
Indeed, research on turn management \citep{sacks1978simplest} has shown that dialogue participants do not start or stop speaking arbitrarily, but follow an underlying turn-taking system to occupy or release the speaker role \citep{petukhova2009s}.
For instance, the last two utterances in Table \ref{table:conversation_sample} illustrate a non-arbitrary change of speakers, following a turn-allocation action (here, a question).
In this conversational situation, speaker B has to take the turn, to respond to speaker A.
To sum up, the sequences of DAs and speakers are tightly interconnected.

However, state-of-the-art DA classification models ignore the sequence of speaker-identifiers \citep{kumar2018dialogue,chen2018dialogue,raheja-tetreault-2019-dialogue,li-etal-2019-dual}.
This is a clear limitation.
To address this limitation, we propose in this paper a simple modification of the CRF layer where the label transition matrix is conditioned on speaker-change.
We evaluate our modified CRF layer within the BiLSTM-CRF architecture, and find that on the SwDA corpus, it improves performance compared to the original CRF.
Furthermore, visualizations demonstrate that sophisticated transition patterns between DA label pairs, conditioned on speaker-change, can be learned in an end-to-end way.

\section{Related work}
In this section, we first introduce the two major DA classification approaches, and then focus on previous work involving the use of BiLSTM-CRF and speaker information.

\noindent\textbf{Multi-class classification}.
In this first approach, consecutive DA labels are considered to be independent.
The DA label of each utterance is predicted in isolation by a classifier such as, e.g., naive Bayes \citep{grau2004dialogue}, Maxent \citep{venkataraman2005does,ang2005automatic}, or SVM \citep{liu2006using}.
Since the first application of neural networks to DA classification by \citet{ries1999hmm}, deep learning has shown promising results even with some simple architectures \citep{khanpour-etal-2016-dialogue,shen2016neural}.
More recent work developed more advanced models, and started taking into account the dependencies among consecutive utterances \citep{kalchbrenner2013recurrent,lee2016sequential,ortega-vu-2017-neural, bothe-etal-2018-context}.
For example, in \citep{bothe-etal-2018-context}, the representations of the current utterance and the three preceding utterances are fed into a RNN, and the last annotation is used to predict the DA label of the current utterance.

\noindent\textbf{Sequence labeling}. In the second approach, the DA labels for all the utterances in the conversation are classified together.
Traditional work uses statistical approaches such as Hidden Markov Models (HMM) \cite{stolcke-etal-2000-dialogue,surendran2006dialog,tavafi-etal-2013-dialogue} and CRFs \citep{lendvai2007token,zimmermann2009joint,kim-etal-2010-classifying} with handcrafted features.
In HMM based approaches, the DA labels are hidden states and utterances are observations emanating from these states. The hidden states are evolving according to a discourse grammar, which essentially is an n-gram language model trained on DA label sequences.
Following advances in deep learning, neural sequence labeling architectures \citep{huang2015bidirectional,reimers2017optimal,yang2018design,cui2019hierarchically} have set new state-of-the-art performance.
Two major architectures have been tested: BiLSTM-Softmax \citep{li-wu-2016-multi,tran-etal-2017-hierarchical,liu-etal-2017-using-context} and BiLSTM-CRF.
This study focuses on the BiLSTM-CRF architecture.

\noindent\textbf{BiLSTM-CRF}.
\citet{kumar2018dialogue} were the first to introduce the BiLSTM-CRF architecture for DA classification.
Their model is hierarchical and consists of two levels, where at level 1, the text of each utterance is separately encoded by a shared bidirectional LSTM (BiLSTM) with last-pooling, resulting in a sequence of vectors.
At level 2, that sequence is passed through another BiLSTM topped by a CRF layer.
At test time, the optimal output label sequence is retrieved from the trained model by Viterbi algorithm \citep{viterbi1967error}.
\citet{chen2018dialogue} and \citet{raheja-tetreault-2019-dialogue} improved on the previous model by adding different attention mechanisms. \citet{li-etal-2019-dual} discovered that performing topic classification as an auxiliary task, can assist in predicting DA labels.
The topic of each utterance is automatically determined using Latent Dirichlet Allocation \citep{blei2003latent}.
Their model consists of two BiLSTM-CRF architectures for predicting simultaneously the target DA label sequence and the target topic label sequence.
This model represents the state-of-the-art in DA classification. 

\noindent\textbf{Speaker information}.
There are only a few previous works that consider the sequence of speaker-identifiers for DA classification.
In \citep{bothe-etal-2018-context}, the utterance representation is the concatenation of the one-hot encoded speaker-identifier, e.g., A as [1, 0] and B as [0, 1], with the output of the RNN-based character-level utterance encoder.
By contrast, \citet{li-wu-2016-multi} and \citet{liu-etal-2017-using-context} choose to concatenate the speaker-change vector with the representation obtained via their CNN-based and RNN-based word-level utterance encoders.
Speaker-change is binary as shown in Table \ref{table:conversation_sample}, obtained by checking if the current utterance is from the same or different speaker as the previous one.
\citet{venkataraman2005does} also include speaker-change as one of the handcrafted features for their Maxent classifier.

Apart from the naive concatenation approaches described above, \citet{kalchbrenner2013recurrent} proposed to let the recurrent and output weights of the RNN cell be conditioned on speaker-identifier, i.e., a speaker-aware RNN cell.
\citet{stolcke-etal-2000-dialogue} proposed to train different discourse grammars for different speakers, to guide DA label transitions in HMM.

\section{Model}\label{sec:model}
\begin{figure}[t]
\centering
\includegraphics[scale=0.31]{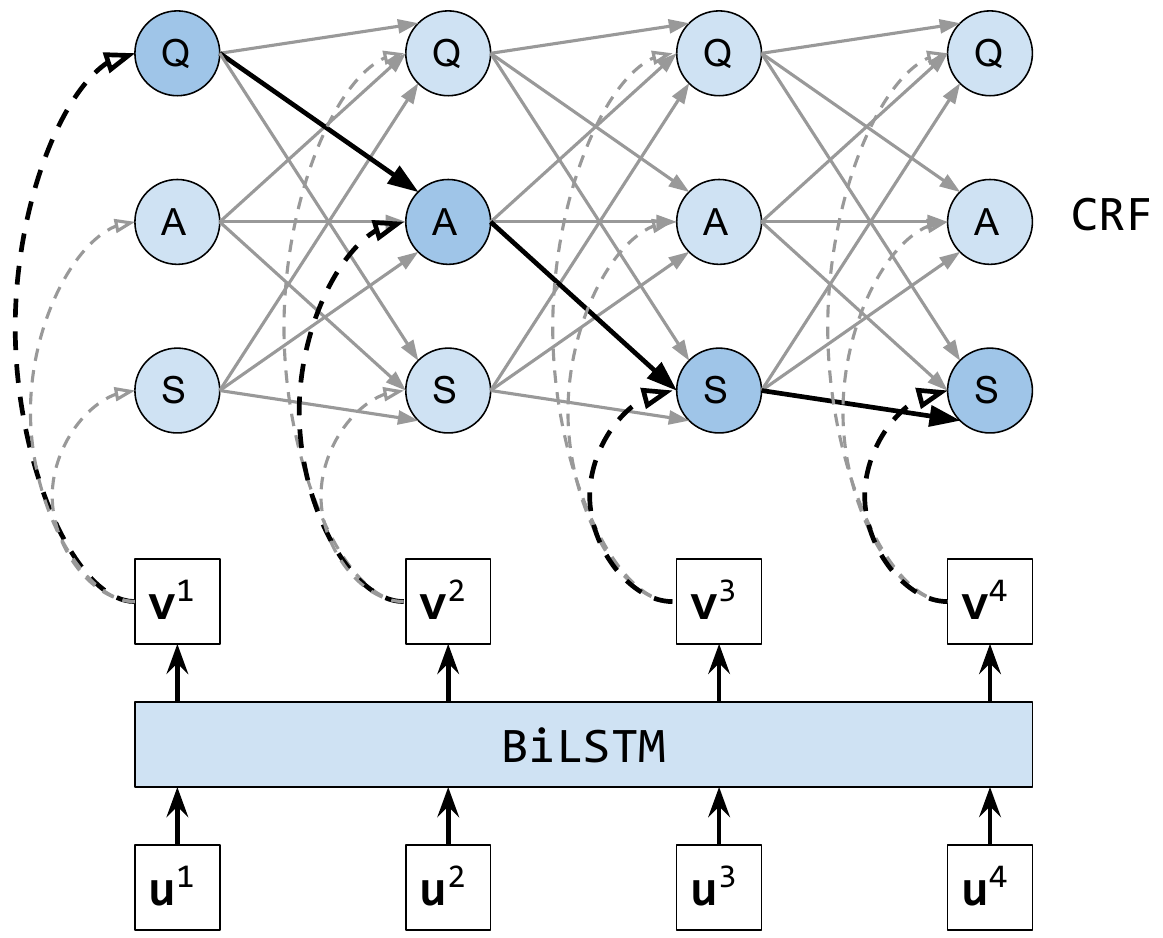}
\caption{
BiLSTM-CRF, for an example with $\{\mathbf{u}^t\}_{t=1}^{T=4}$ (utterance embeddings) as input and $Q,A,S,S$ (DA labels) as target. Three possible labels $\{Q,A,S\}$ stand for Question, Answer, and Statement, respectively.
}
\label{fig:BiLSTM-CRF}
\end{figure}

Here, we describe the general BiLSTM-CRF model for DA classification, shown in Fig. \ref{fig:BiLSTM-CRF}. Then, in the next section, we present our modification of the CRF layer that takes speaker-change into account.

\noindent\textbf{Notation}.
Let us denote by $\{(\mathbf{x}^t, y^t)\}_{t=1}^{T}$ a conversation of length $T$.  $X=\{\mathbf{x}^t\}_{t=1}^{T}$ is the sequence of utterances, where each utterance $\mathbf{x}^t=\{x_n^t\}_{n=1}^N$ is itself a sequence of words of length $N$.
$Y=\{y^t\}_{t=1}^{T}$ denotes the target sequence, where $y^t \in \mathcal{Y}$ is the set of all possible DA labels of size $|\mathcal{Y}|=K$.
We use $y^t$ to denote the label and its integer index interchangeably.

\noindent\textbf{Utterance encoder}.
Each utterance is separately encoded by a shared forward RNN with LSTM cells.
Only the last annotation $\mathbf{u}^t_N$ is retained (last pooling).
We are left with a sequence of utterance embeddings $\{\mathbf{u}^t\}_{t=1}^T$.

\noindent\textbf{BiLSTM layer}. 
The sequence of utterance embeddings $\{\mathbf{u}^t\}_{t=1}^T$ is then passed on to a bidirectional LSTM, returning the sequence of conversation-level utterance representations $\{\mathbf{v}^t\}_{t=1}^T$.

\noindent\textbf{CRF layer}.
$\mathbf{v}^t$ can already be used to predict locally the label at each time step in isolation, through a dense layer with softmax activation, which results in the BiLSTM-Softmax architecture.
However this might lead to a non-optimal global solution, if we consider the output DA label sequence as a whole.

On the other hand, CRF models the conditional probability $P(Y|X)$ of an entire target sequence $Y$ given an entire input sequence $X$.
Thus, it guarantees an optimal global solution, under the first order Markov assumption.
More precisely:

{\small
\setlength{\abovedisplayskip}{-3pt}
\setlength{\belowdisplayskip}{3pt}
\begin{align}
\label{eq:main}
    P(Y|X) = \frac{\exp(\psi(X,Y))}{\sum_{\tilde{Y}} \exp(\psi(X,\tilde{Y}))}
\end{align}
}

\noindent where $\psi(X,Y)$ is a feature function that assigns a \textit{path score} to the label sequence $Y$, giving $X$.
Then, a softmax function is used to yield the conditional probability, where $\tilde{Y}$ denotes one of all possible label sequences (paths).

$\psi(X,Y)$ is defined as the sum of \textit{emission scores} (or state scores) and \textit{transition scores} over all time steps \citep{morris2006combining,chen2019transfer}:

{\small
\setlength{\abovedisplayskip}{-3pt}
\setlength{\belowdisplayskip}{3pt}
\begin{align}
    \psi(X,Y) = \sum_{t=1}^{T} h(y^t, X) + \sum_{t=1}^{T-1} g(y^t,y^{t+1})
\end{align}
}

\noindent Emission (state) scores are assigned to the dashed top-down edges (nodes) in Fig. \ref{fig:BiLSTM-CRF}, computed as follows:

{\small
\setlength{\abovedisplayskip}{-3pt}
\setlength{\belowdisplayskip}{3pt}
\begin{align}
    h(y^t, X) = (\mathbf{W}\mathbf{v}^t + \mathbf{b})[y^t]
\end{align}
}

\noindent where the conversation-level utterance representation $\mathbf{v}^t$ is converted into a vector of size $K$ and $[y^t]$ denotes the element at index $y^t$.
Higher values of $h(y^t, X)$ indicate that the model is more confident in predicting the output label $y^t$ at time step $t$.

\noindent Transition scores are assigned to the solid left-to-right edges in Fig. \ref{fig:BiLSTM-CRF}, computed as follows:

{\small
\setlength{\abovedisplayskip}{-3pt}
\setlength{\belowdisplayskip}{3pt}
\begin{align}
    g(y^t,y^{t+1}) = \mathbf{G}[y^t,y^{t+1}]
\end{align}
}

\noindent where $\mathbf{G}$ is the label transition matrix of size $K \times K$.
E.g, the element $\mathbf{G}[y^t,y^{t+1}]$ is the transition score from label $y^t$ to label $y^{t+1}$.
Note that the transition matrix is shared across all time steps.

The CRF layer is parameterized by $\mathbf{W}$, $\mathbf{b}$, and $\mathbf{G}$.
To learn these parameters and those of the previous layers, maximum likelihood estimation is used.
For a training set of $M$ conversations, the loss can be written as:

{\small
\setlength{\abovedisplayskip}{-3pt}
\setlength{\belowdisplayskip}{3pt}
\begin{align}
   \mathcal{L} = \sum_{m=1}^M - \log P(Y^{m}|X^{m}) 
\end{align}
}

\noindent At test time, the optimal output label sequence, i.e., $Y^* = \operatorname*{argmax}_{\tilde{Y}} P(\tilde{Y}|X)$ for unseen $X$, is obtained with the Viterbi decoding algorithm \citep{viterbi1967error}.
Due to the Markov property of the linear chain CRF, the computations of Viterbi algorithm and the normalization term in Eq. \ref{eq:main} can be broken down into a series of sub-problems over time in a recursive manner, which are solved via dynamic programming \citep{bellman1966dynamic}, with polynomial complexity $\mathcal{O}(TK^2)$.

\section{Our contribution}
Given the sequence of speaker-identifiers $S=\{s^t\}_{t=1}^{T}$, we can instantly derive the sequence of speaker-changes $Z=\{z^{t, t+1}\}_{t=1}^{T-1}$ by comparing neighbors. E.g., $z^{2,3}=0$ means the speaker does not change from time $t=2$ to $t=3$. 

We extend the original CRF so that it considers as additional input, the sequence $Z$.
That is, CRF now models $P(Y|X,Z)$ instead of just $P(Y|X)$.
In other words, the prediction of the DA label sequence is now conditioned both on the utterance sequence and the speaker-change sequence.
Specifically, transition scores in our modified CRF layer are computed as follows:

{\small
\setlength{\abovedisplayskip}{-3pt}
\setlength{\belowdisplayskip}{3pt}
\begin{align}
    g(y^t,y^{t+1},z^{t,t+1}) =  &(1-z^{t,t+1})*\mathbf{G}_0[y^t,y^{t+1}] + \nonumber \\
    &z^{t,t+1}*\mathbf{G}_1[y^t,y^{t+1}]
\end{align}
}

\noindent where $\mathbf{G}_0$ and $\mathbf{G}_1$ are label transition matrices of size $K \times K$, corresponding respectively to the ``speaker unchanged'' and ``speaker changed'' cases.

\section{Experimental setup}

\begin{figure}[t]
\centering
\includegraphics[scale=0.5]{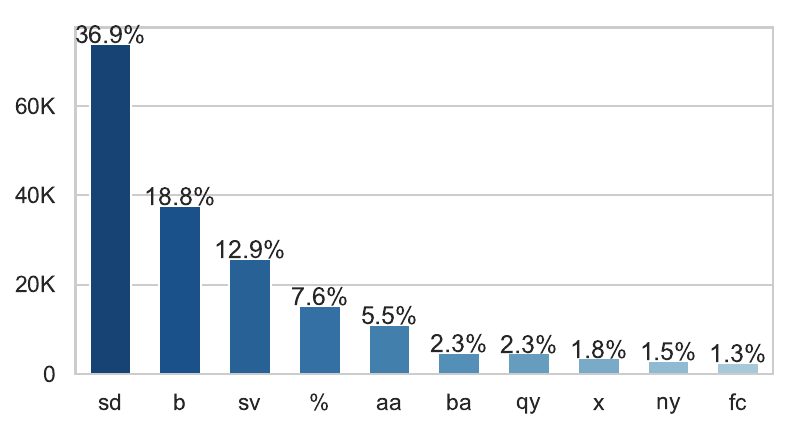}
\caption{Counts and frequencies of the 10 most represented DA labels in the SwDA dataset.
There are 200444 utterances in total.}
\label{fig:barplot}
\end{figure}

\noindent \textbf{Dataset}.
We experiment on the widely-used SwDA\footnote{\url{https://github.com/cgpotts/swda}} (Switchboard Dialogue Act) dataset \citep{jurafsky1997switchboard,stolcke-etal-2000-dialogue}.
This corpus contains telephone conversations recorded between two randomly selected speakers talking about one of various general topics (air pollution, music, football, etc.).
In this dataset, utterances are annotated with 42 mutually exclusive DA labels, based on the SWBD-DAMSL annotation scheme \citep{jurafsky1997switchboard}.
Inter-annotator agreement is 84\%.
The frequency of the 10 most represented DA labels are illustrated in Fig. \ref{fig:barplot}. 
We can see that labels are highly imbalanced and follow a long-tailed distribution.
Detailed statistics for all 42 labels are provided in Appendix \ref{app:da_label_statistics}.

We adopt the same training, validation and testing partition as previous work \citep{lee2016sequential}\footnote{\url{https://github.com/Franck-Dernoncourt/naacl2016}}, consisting of 1003, 112, and 19 conversations, respectively.

\noindent \textbf{A note about the `+' tag}.
The `+' tag, as shown in Table \ref{table:conversation_sample}, accounts for 8.1\% of the total annotations, but is not part of the default label set.
That tag is used to mark the remaining parts of an utterance that has been interrupted by the other speaker.
While most of the previous works did not predict, or even mention this tag, some efforts considered it as a 43$^{rd}$ DA label and predicted it \citep{lee2016sequential,raheja-tetreault-2019-dialogue}.

In this paper, we followed the approach of \citep{webb2005dialogue,milajevs-purver-2014-investigating,kim2017compositional}, and used the `+' tag to reconnect, bottom-up, all the parts of an interrupted utterance together.
E.g., in Table \ref{table:conversation_sample}, the parts ``{\small \texttt{Of course I use,}}'' and ``{\small \texttt{credit cards.}}'', uttered by speaker B, are reconnected into ``{\small \texttt{Of course I use, credit cards.}}'', which becomes the new first utterance.
It is followed by ``{\small \texttt{<laughter>}}'', uttered by speaker A.
We opted for this approach as predicting the DA of a broken utterance sometimes does not make sense.
For instance, in this situation with three utterances: (1) ``{\small \texttt{A: so, (Wh-Question)}}'', (2) ``{\small \texttt{B: <throat\_clearing> (Non-verbal)}}'', and (3) ``{\small \texttt{A: what's your name? (+)}}'', it is very difficult to correctly predict that utterance 1 is a question.
And predicting anything other than a question-related tag for utterance 3 does not really make sense.
Reconstructing 1 and 3 into a single utterance ``{\small \texttt{A: so, what's your name? (Wh-Question)}}'' solves both issues.

\noindent\textbf{Implementation and training details}.
Disfluency markers \citep{meteer1995dysfluency} were filtered out and all characters converted to lowercase.
We used some optimal hyperparameters provided by \citet{kumar2018dialogue}.
E.g., 0.2 dropout was applied to the utterance embeddings and conversation-level utterance representations, and all LSTM layers had 300 hidden units.
The embedding layer was initialized using 300-dimensional word vectors pre-trained with the gensim \cite{rehurek_lrec} implementation of {\small\texttt{word2vec}} \citep{mikolov2013efficient} on the utterances of the training set, and was frozen during training.
Vocabulary size was around 21K, and out-of-vocabulary words were mapped to a special token [UNK], randomly initialized. 

Models were trained with the Adam optimizer \citep{kingma2014adam}.
Early stopping was used on the validation set with a patience of 5 epochs and a maximum number of epochs of 100.
The best epoch was selected as the one associated with the highest validation accuracy. 
Usually, the best epoch was within the first 10.
We set our batch-size to be 1, i.e, one conversation for one training step.
Batch sizes of 1, 2, 4, 8, and 16 were also tried, without observing significant differences.

\begin{figure*}[ht]
\centering
\begin{subfigure}[t]{0.49\textwidth}
\centering
\includegraphics[scale=0.6]{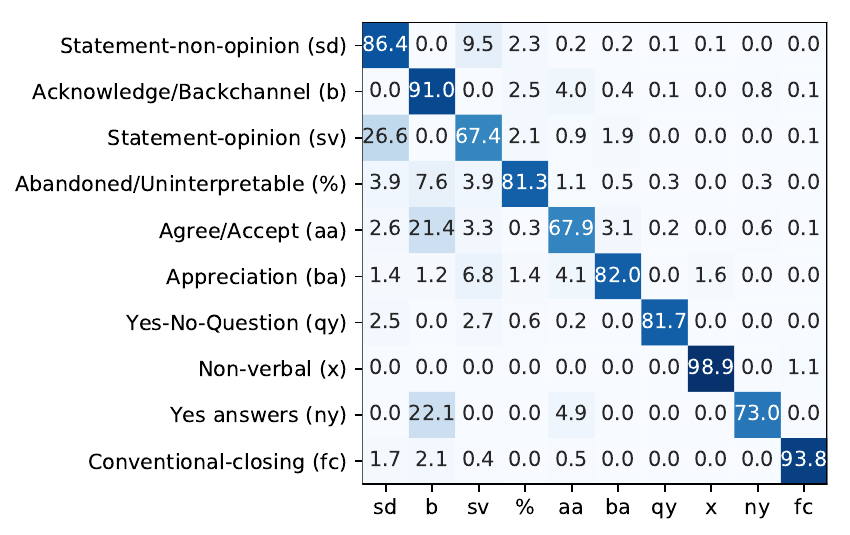}
\end{subfigure}
\begin{subfigure}[t]{0.49\textwidth}
\centering
\includegraphics[scale=0.6]{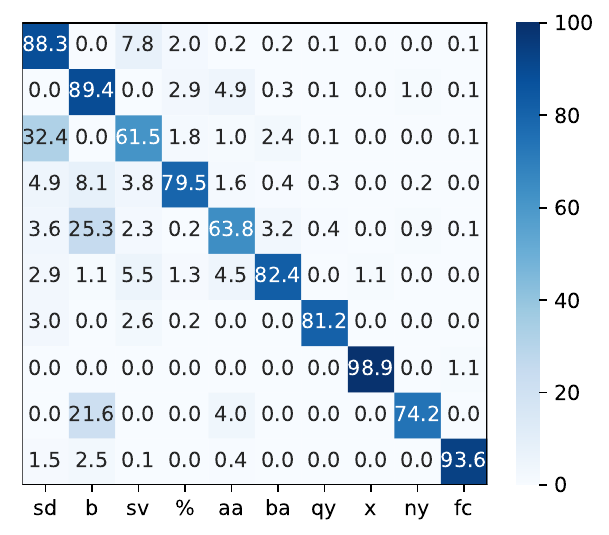}
\end{subfigure}
\caption{Normalized confusion matrices, averaged over 10 runs, for the 10 most frequent DA labels (90.9\% of all annotations).
Left: our model, right: base model.
Rows (columns) correspond to true (predicted) classes.}
\label{fig:confusion}
\end{figure*}

\section{Quantitative results}
\noindent \textbf{Performance comparison}. Table \ref{table:results} reports the results in terms of classification accuracy, averaged over 10 runs to account for the randomness of SGD.
Model a) uses our modified CRF layer. Model b) has exactly the same architecture as a), but uses a vanilla CRF layer.

Results show that, in terms of overall accuracy on the test set of 42 DA labels, our model a) outperforms the base model b) by 1\%.
Moreover, the small standard deviations highlight the consistency of this improvement over the 10 runs.
Note that this performance gain is solely caused by our modified CRF layer capturing speaker-change, and is greater than the gains of 0.26\% \citep{liu-etal-2017-using-context} and 0.09\% \citep{bothe-etal-2018-context} reported by previous attempts at leveraging speaker information.

To interpret the results in more detail, we show in Fig. \ref{fig:confusion} the confusion matrices of our model and the base model, for the 10 most frequent DA labels, representing close to 91\% of all annotations.
The rows correspond to true classes, and the columns to predicted classes.\footnote{Note that our confusion matrices were row-wise normalized by class size. So we use the terms accuracy (per class) to denote diagonal values (equivalent to recall or hit rate), and miss rate for off-diagonal values.}
By looking at the diagonals, we can see that our model (on the left) better predicts 6 labels out of 10 with absolute accuracy gains of up to 5.9\% (for statement-opinion, \texttt{sv})\footnote{The margins are even larger (up to 20\%) on some less frequent labels, as shown by the results in Appendix \ref{app:case_analysis}.} and is on par with the baseline model for one label (non-verbal, \texttt{x}), at 98.9\%.
By looking at off-diagonal values, miss rates are decreased up to 5.8\% (for \texttt{sv} misclassified as \texttt{sd}) by our model.
Also, our model provides a large boost for the (Acknowledge/Backchannel, Agree/Accept), or (\texttt{b}, \texttt{aa}) pair.
It increases the respective accuracies by 1.6\% (89.4\%$\rightarrow$91.0\%) and 4.1\% (63.8\%$\rightarrow$67.9\%).
The respective miss rates are decreased by 0.9\% (4.9\%$\rightarrow$4.0\%) and 3.9\% (25.3\%$\rightarrow$21.4\%), respectively for \texttt{b} misclassified as \texttt{aa} and  \texttt{aa} misclassified as \texttt{b}.
This is to be noted, as these two labels are among the most frequently confused pairs \citep{kumar2018dialogue,chen2018dialogue}.

\begin{table}[t]
\small
\centering
\setlength{\tabcolsep}{4pt} 
\renewcommand{\arraystretch}{1.2} 
\scalebox{1}{
\begin{tabular}{|rl|l|c|c|}
\hline
&  & BiLSTM  & CRF      & Accuracy \\ 
& Model & ~~~input & extra input & (\% $\pm$ SD) \\ 
\hline
a)&     Our CRF & $\mathbf{u}^t$ & SC & \textbf{78.70} $\pm$ .37 \\
a1)&            & $\mathbf{u}^t$ + SI & SC & 78.32 $\pm$ .28 \\
a2)&            & $\mathbf{u}^t$ + SC & SC & 78.65 $\pm$ .47 \\
\hline
b)&     Vanilla CRF & $\mathbf{u}^t$      & - & 77.69 $\pm$ .38 \\
b1)&                & $\mathbf{u}^t$ + SI & - & 77.86 $\pm$ .61 \\
b2)&                & $\mathbf{u}^t$ + SC & - & 78.33 $\pm$ .71 \\
\hline
c)&     Softmax & $\mathbf{u}^t$      & - & 77.80 $\pm$ .48 \\
c1)&            & $\mathbf{u}^t$ + SI & - & 77.73 $\pm$ .44 \\
c2)&            & $\mathbf{u}^t$ + SC & - & 78.33 $\pm$ .49 \\
\hline
\multicolumn{2}{|c|}{a) + b)} & \multirow{2}{*}{$\mathbf{u}^t$} & \multirow{2}{*}{SC} & \multirow{2}{*}{\textbf{78.89} $\pm$ .20} \\
\multicolumn{2}{|c|}{ensembling} & & &\\
\hline
\multicolumn{2}{|c|}{a) + b)} & \multirow{2}{*}{$\mathbf{u}^t$} & \multirow{2}{*}{SC} & \multirow{2}{*}{78.27 $\pm$ .47} \\
\multicolumn{2}{|c|}{joint training} & & & \\
\hline
\end{tabular}
}
\caption{Results, averaged over 10 runs. SI: speaker-identifier, SC: speaker-change, $\mathbf{u}^t$: utterance embedding, $\pm$: standard deviation.}
\label{table:results}
\end{table}

\begin{table}[ht]
\small
\centering
\setlength{\tabcolsep}{5.5pt} 
\renewcommand{\arraystretch}{1.2} 
\scalebox{1}{
\begin{tabular}{|r|c|ccc|}
\hline
\multicolumn{2}{|c|}{} & P & R & F1 \\
\hline
Our &sd & 80.49 & 86.36 & 83.32 \\
&sv & 71.54 & 67.41 & 69.42 \\
\hline
Vanilla &sd & 77.83 & 88.32 & 82.74 \\
&sv & 73.24 & 61.48 & 66.84 \\
\hline
\end{tabular}
}
\caption{Precison, Recall, and F1 score (\%) of our model vs. base model on the \texttt{sd} and \texttt{sv} labels.}
\label{table:sd_sv}
\end{table}

Although our model achieves significant gains on a majority of the most frequent labels, it decreases performance for the most frequent label, \texttt{sd}, which accounts for 36.9\% of all labels, as shown by Fig. \ref{fig:barplot}.
This explains why, in terms of overall accuracy, our improvements are modest.
In addition, the performance drop regarding \texttt{sd} can be interpreted as a consequence of the trade-off between \texttt{sd} and \texttt{sv}, since the distinction between them was very hard to make even by annotators \citep{jurafsky1997switchboard}.
This can be demonstrated in terms of precision, recall, and F1 score, as shown in Table \ref{table:sd_sv}.
We can observe that, as opposed to the base model, our model has lower \texttt{sd} and higher \texttt{sv} recall values.
A similar observation can be made for precision scores.
Thus, the prediction between \texttt{sd} and \texttt{sv} is a trade-off made by models.
It is also interesting to note that our model is superior for both labels in terms of F1 score.

\begin{table}[ht]
\small
\centering
\setlength{\tabcolsep}{4pt} 
\renewcommand{\arraystretch}{1.2} 
\scalebox{1}{
\begin{tabular}{|l|l|l|l|}
\hline
&  Ours & Vanilla & Diff. \\
\hline
10 best DAs & 37.08 & 31.70 & + 5.38\\
\hline
10 worst DAs & 59.67 & 64.54 & - 4.87\\
\hline
\end{tabular}
}
\caption{Average accuracy (\%) of our model vs. base model on the 10 DAs best and worst predicted by our model (resp. representing 20\% and 40\% of all annotations).}
\label{table:results_case_analysis}
\end{table}

We can observe in Table \ref{table:results_case_analysis} that our model brings improvement where it is most necessary, i.e., for the most difficult and rare DAs (20\%).
Full details are provided in Appendix \ref{app:case_analysis}, along with the corresponding confusion matrices.

\noindent \textbf{The benefits of considering speaker information vary across DA labels}.
Our model and the base model performed very closely on 4 labels: Non-verbal (\texttt{x}), Conventional-closing (\texttt{fc}), Appreciation (\texttt{ba}), and Yes-No-Question (\texttt{qy}).
We found that the utterances of these labels contain clear lexical cues that can be mapped to corresponding DA labels in a \textit{non-ambiguous} way.
Some examples include ``{\small \texttt{<laughter>}}'' $\rightarrow$ \texttt{x},  ``{\small \texttt{Bye-bye.}}'' $\rightarrow$ \texttt{fc},  ``{\small \texttt{That's great.}}'' $\rightarrow$ \texttt{ba}, and ``{\small \texttt{Do you \dots ?}}'' $\rightarrow$ \texttt{qy}. 
In other words, predicting well these four DAs does not require having access to speaker information.
It can be done solely from the text of the current utterance.
Having access to context is not even required.
This explain why our speaker-aware CRF is not helpful here.

This interpretation is supported by the fact that, as explained in Appendix \ref{app:case_analysis}, our model is most useful for the DAs that require speaker-change awareness.

\noindent \textbf{Ensembling and joint training}.
Since the model using our CRF and the model using the vanilla CRF appear to have their own strengths and weaknesses, we tried combining them to improve performance.
More precisely, we experimented with two approaches.
First, an ensembling approach that combines the predictions of the two trained models by averaging their emission and transition scores (respectively).
Second, a joint training approach that combines the two models into a new one and trains it from scratch.
In that second model, our CRF and the vanilla CRF are combined, and transition scores are computed as:

{\small
\setlength{\abovedisplayskip}{-3pt}
\setlength{\belowdisplayskip}{3pt}
\begin{align}
    &g(y^t,y^{t+1},z^{t,t+1}) = \mathbf{G}_{basis}[y^t,y^{t+1}] +  \\
    &(1-z^{t,t+1})*\mathbf{G}_0[y^t,y^{t+1}] + z^{t,t+1}*\mathbf{G}_1[y^t,y^{t+1}] \nonumber
\end{align}
}

\noindent where ${G}_{basis}$ is the transition matrix as in the original CRF, used at each time step, while $\mathbf{G}_0$/$\mathbf{G}_1$ are applied only when the speaker does not change/changes, as in our modified CRF layer.

Results in Table \ref{table:results} show that the ensemble model reaches new best performance at 78.89, providing close to a 0.2 boost from our model, and a 1.2 boost from the vanilla CRF model.
On the other hand, the jointly-trained model does not outperform our model.
After inspecting the transition matrices for the two cases ($\mathbf{G}_{basis}+\mathbf{G}_{0}$) and ($\mathbf{G}_{basis}+\mathbf{G}_{1}$), we found that the addition of $\mathbf{G}_{basis}$ blurred the label transition patterns.

\begin{figure*}[ht]

\centering
\includegraphics[width=.427\textwidth,trim={0 0 0.25cm 0},clip]{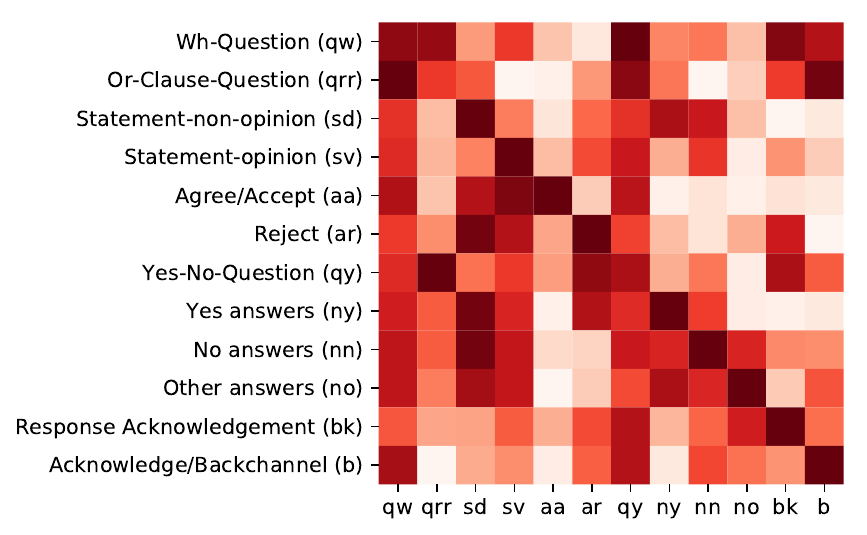}
\includegraphics[width=.245\textwidth,trim={0 0 2cm 0},clip]{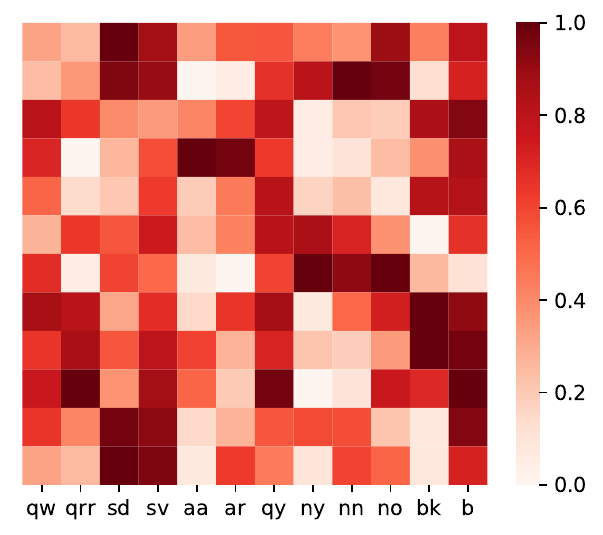} \hspace{0.15cm}
\includegraphics[width=.29\textwidth,trim={1.5cm 0 0 0},clip]{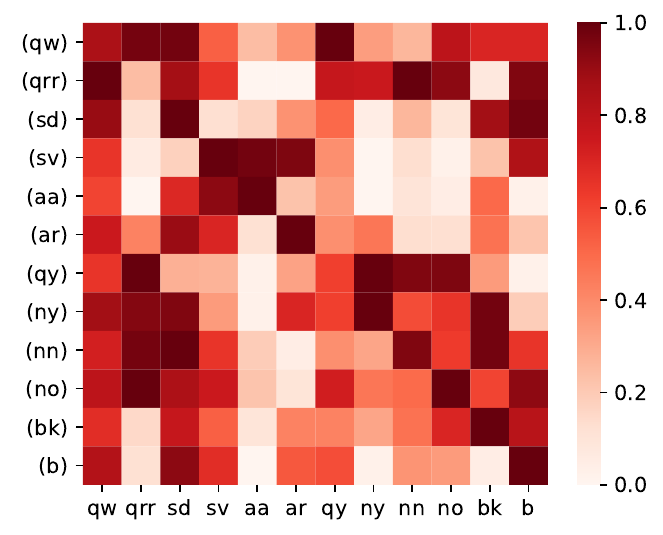}

\caption{Normalized transition matrices (averaged over 10 runs). Left and center: $\mathbf{G}_0$ (speaker unchanged) and $\mathbf{G}_1$ (speaker changed) of our CRF layer. Right: $\mathbf{G}$ of vanilla CRF layer.
The darker, the greater the score. \label{fig:transition}}
\end{figure*}

\noindent \textbf{Ablation studies}.
Our results showed that considering speaker information improves DA classification.
Then, we wanted to confirm whether our way of taking speaker information into account (at the CRF level) was the most effective.
To this purpose, we trained two other base models, both using the vanilla CRF.
These two models respectively concatenate the one-hot encoded speaker-identifier vector (noted SI, of size 2) and the binary speaker-change vector (noted SC, of size 1) with the utterance embedding $\mathbf{u}^t$.\footnote{proposed in \citep{liu-etal-2017-using-context,bothe-etal-2018-context}, but not in the context of BiLSTM-CRF.}
Results, shown in rows b1 and b2 of Table \ref{table:results}, show that while they bring improvement compared to the basic base model (row b), these two approaches are not able to yield as big of a gain as the model using our modified CRF layer, indicating that taking speaker-change into account at the CRF level is superior.

For the sake of completeness, we repeated these experiments with our model.
Results, available in rows a1 and a2 of Table \ref{table:results}, show that performance was not improved (78.32 and 78.65 vs. 78.70).
Thus, it seems that taking speaker information into account twice, both at the BiLSTM level and at the CRF level, is not useful, or at least, not in this way.

Results in Table \ref{table:results} also show that SC is a better feature than SI in general.

\noindent \textbf{BiLSTM-CRF vs. BiLSTM-Softmax}.
To the best of our knowledge, no previous study has compared BiLSTM-CRF to BiLSTM-Softmax on the DA classification task.
Hence, in this paper, we decided to compare between these two models.
Results reveal that the models using BiLSTM-Softmax (rows c, c1, and c2) are competitive with the ones using BiLSTM-CRF (rows b, b1, and b2).
More specifically, BiLSTM-Softmax outperforms BiLSTM-CRF with text features only (rows b vs. c), by a slight 0.11 margin, but it is the opposite for text + SI (b1 vs. c1, 0.13 difference).
With text + SC (b2 vs. c2), they achieve similar performance.

These results are not very surprising, since, on other tasks than DA classification,  multiple recent works have reported that BiLSTM-CRF does not always outperform  BiLSTM-Softmax \citep{reimers2017optimal,yang2018design,cui2019hierarchically}.
For example, in \cite{yang2018design}, CRF brought improvement for named entity recognition and chunking, but not for POS tagging. 
One of the reasons might be that the simple Markov label transition model of CRF does not give much information gain over strong neural encoding \citep{cui2019hierarchically}.
That is, BiLSTM may be expressive enough to implicitly capture the obvious dependencies among labels.

In any case, the model equipped with our CRF layer (row a) outperforms all variants of BiLSTM-Softmax and BiLSTM-vanilla\_CRF.
This suggests that our CRF layer can capture richer and \textit{not obvious} label dependencies given speaker information, which, in the end, makes the use of a CRF layer valuable in assisting DA classification.

\section{Qualitative results}
\noindent \textbf{Visualization of transition matrices}.
We illustrate, in Fig. \ref{fig:transition}, the transition matrices $\mathbf{G}_0$ and $\mathbf{G}_1$ of our CRF layer, together with the single matrix $\mathbf{G}$ of the vanilla CRF layer.
This visualization is done for 12 labels that are easy to interpret, such as statements, questions, answers, etc. 
We can observe some interesting patterns, sometimes matching intuition, and sometimes harder to interpret.
We report some of the most interesting findings below:

\noindent \textbf{1)} Overall, $\mathbf{G}_0$ and $\mathbf{G}_1$ are not identical, which means that different transition patterns are associated with the ``speaker unchanged'' and ``speaker changed'' cases. 
The dark diagonal of $\mathbf{G}_0$ shows that when the speaker does not change, the majority of labels tend to carry over to the next utterance.
On the opposite, $\mathbf{G}_1$ clearly shows that changing speakers very often induce a change in DA.

\noindent \textbf{2)} questions starting with words including: `what', `how', etc. (\texttt{qw} label) tend to transfer to statements (\texttt{sd} and \texttt{sv}) and to other answers (\texttt{no}, e.g., ``{\small\texttt{I don't know}}'') if the speaker changed, but to other forms of questions, yes-no questions and questions starting with the word `or' (\texttt{qy} and \texttt{qrr}), or to acknowledgements (\texttt{bk} and \texttt{b}) if the speaker did not change.
This probably corresponds to instances when the same speaker clarifies, elaborates on, or answers, an original question.

\noindent \textbf{3)} \texttt{sv} label (statement with opinion) tends to transition to Agree/Accept \texttt{aa} and Reject \texttt{ar} if the speaker changed, while no such clear pattern can be observed for the \texttt{sd} label (statement without opinion).

\noindent \textbf{4)} \texttt{qy} label (yes/no questions) tend to transfer to answer labels \texttt{ny} (yes), \texttt{nn} (no), \texttt{no} (other) if the speaker changed, and to another type of question (e.g., or-clause) if the speaker did not change.
Again, the latter surely corresponds to the case where a given speaker elaborates on his or her original question.

\noindent \textbf{5)} answer labels (\texttt{ny}, \texttt{nn}, \texttt{no}) tend to be followed by Response Acknowledgement \texttt{bk} and Acknowledge/Backchannel \texttt{b} if the speaker changed, but by themselves or statements (\texttt{sd} and \texttt{sv}) if the speaker did not change.

As far as the transition matrix $\mathbf{G}$ of the vanilla CRF layer (right of Fig. \ref{fig:transition}), we can observe that it tries to capture, at the same time, the transition patterns of both the ``speaker changed'' and ``speaker unchanged'' cases.
For example, \texttt{sv} equally tends to transfer to \texttt{sv}, \texttt{aa} and \texttt{ar} in $\mathbf{G}$, while the transitions towards \texttt{sv}/\texttt{aa},\texttt{ar} are only probable if the speaker stays the same/changes, as clearly illustrated by $\mathbf{G_0}$/$\mathbf{G_1}$.
Obviously, using two matrices as in our approach gives much more expressiveness to the model in capturing DA label transition patterns.
To summarize, visualizations show that the transition matrices $\mathbf{G}_0$ and $\mathbf{G}_1$ in our modified CRF layer are able to encode speaker-change-aware, sophisticated DA transition patterns.

\section{Discussion}

Note that, for our utterance encoder, we also experimented with a bidirectional LSTM (also with last pooling), as in \citep{kumar2018dialogue}, and with a bidirectional LSTM with self-attention mechanism \citep{yang-etal-2016-hierarchical}.
However, since they were not giving better results, we opted for the simplest option.
One possible explanation for the self-attention mechanism not being helpful could be the very short size of the utterances in the SwDA dataset (68.7\% of utterances are shorter than 10 tokens).
On such short sequences, a RNN with a 300-dimensional hidden layer is very likely able to keep the full sequence into memory.
As far as why a forward RNN suffices, it should be noted that with last pooling, the last time step corresponds to the first annotation of the backward RNN.
This is not adding much information to the last annotation of the forward RNN, which represents the entire sequence.

Our goal was not to exceed the state-of-the-art accuracy reported in \cite{li-etal-2019-dual,raheja-tetreault-2019-dialogue}, this is why we used simple models in all of our experiments.
However, our improved CRF layer can be directly plugged into more advanced architectures, such as Att-BiLSTM-CRF \citep{luo2018attention} or Transformer-CRF \citep{chen2019bert,zhang2019using,yan2019tener,winata2019hierarchical}, and should in principle be able to boost performance regardless of the model used.

Future research should be devoted to address the limitation of the Markov property of CRF layer, by developing a model that is capable of capturing longer-range dependencies within and among the three sequences: that of speakers, utterances, and DA labels.

\section{Conclusion}
In this paper, we focused on demonstrating that taking speaker information into consideration was beneficial to the task of DA classification, with the BiLSTM-CRF architecture.
We proposed a modified CRF layer that takes as extra input the sequence of speaker-changes. Experiments conducted on the SwDA dataset showed that our CRF layer outperforms vanilla CRF, and brings greater gains than previous attempts at taking speaker information into account.
Moreover, visualizations confirmed that our improved CRF was able to learn complex speaker-change aware DA transition patterns in an end-to-end way.

\section{Acknowledgments}
This work was supported by the \href{https://linto.ai}{LinTo} project \citep{lorre2019linto}.

\onecolumn

\begin{center}
\textbf{\Large Speaker-change Aware CRF for Dialogue Act Classification}
\vspace{0.2cm}

\textbf{\Large \textit{Supplementary Material}}
\vspace{0.6cm}
\end{center}

\noindent\textbf{\Large Appendices}

\appendix

\section{Worst and best case analysis}\label{app:case_analysis}
\begin{figure*}[ht]
\centering
\begin{subfigure}[t]{0.49\textwidth}
\hspace{0.7em}
\includegraphics[scale=0.6]{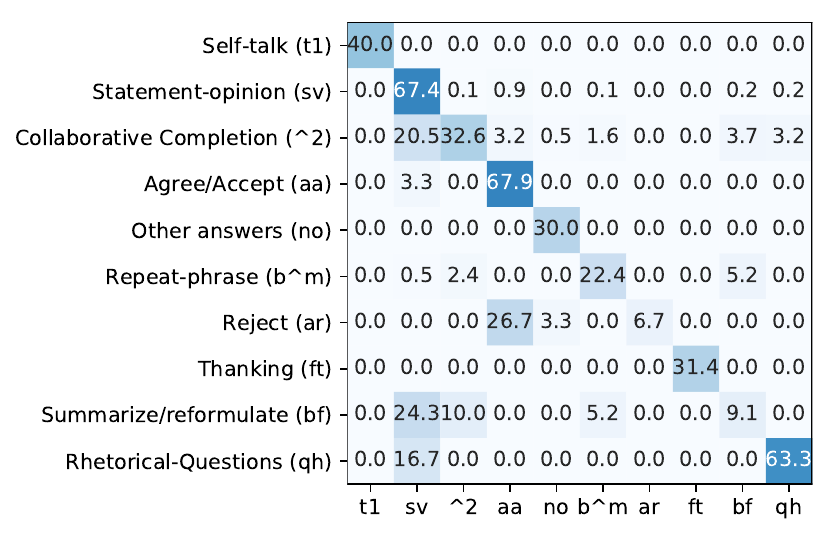}
\end{subfigure}
\begin{subfigure}[t]{0.49\textwidth}
\centering
\includegraphics[scale=0.6]{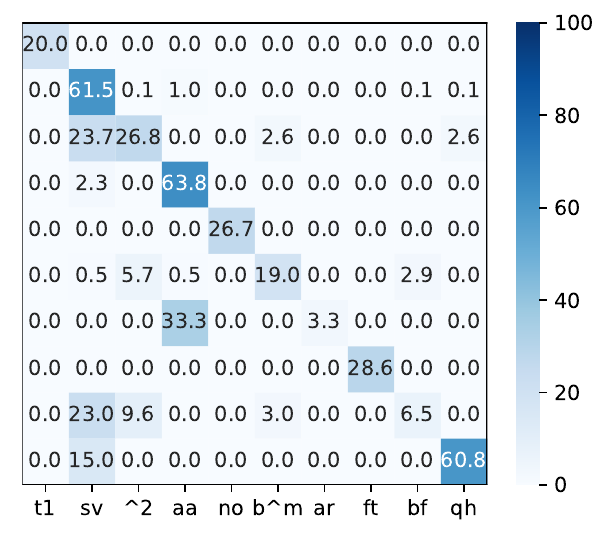}
\end{subfigure}
\caption{Normalized confusion matrices, averaged over 10 runs, for the 10 DA labels \textbf{best} predicted by our model (20.2\% of all annotations). Left: our model, right: base model.}
\label{fig:best_confusion}
\end{figure*}

\begin{figure*}[ht]
\centering
\begin{subfigure}[t]{0.49\textwidth}
\hspace{-0.7em}
\includegraphics[scale=0.6]{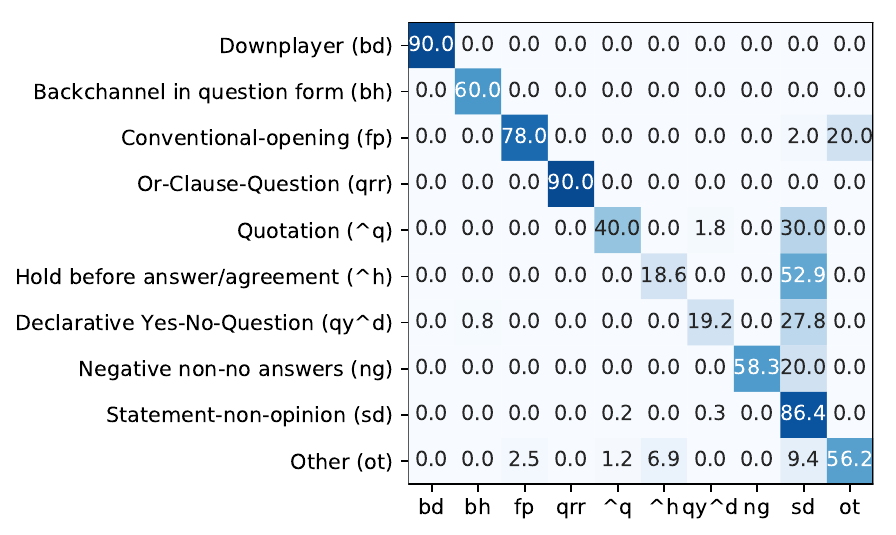}
\end{subfigure}
\begin{subfigure}[t]{0.49\textwidth}
\centering
\includegraphics[scale=0.6]{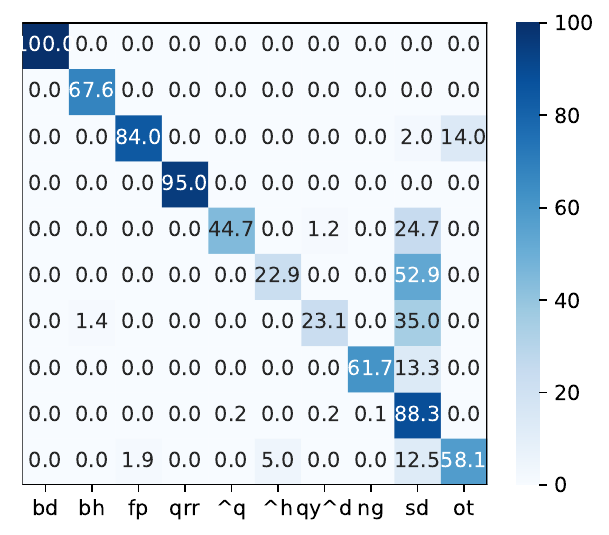}
\end{subfigure}
\caption{Normalized confusion matrices, averaged over 10 runs, for the 10 DA labels \textbf{worst} predicted by our model (39.6\% of all annotations). Left: our model, right: base model.}
\label{fig:worst_confusion}
\end{figure*}

In this section, we interpret the confusion matrices for the 10 DA labels that our model best predicted (Fig. \ref{fig:best_confusion}) and worst predicted (Fig. \ref{fig:worst_confusion}), in comparison with the base model, always on the right.
Inspecting the matrices reveals that our model is most useful for the DAs requiring speaker-change awareness, which confirms the effectiveness of our modification of the CRF layer.
It also shows that our model brings improvement where it is most necessary, i.e., for the most difficult and rare DAs.\\

\noindent \textbf{Relative differences}. For the 10 DA labels best predicted by our model, the average performance gain compared to the base model is equal to \textbf{5.38} (shown in Table \ref{table:results_case_analysis}), whereas the drop in performance for the 10 DAs worst predicted by our model is lower, only equal to \textbf{4.87}.
Thus, when it improves performance, our model does so with a greater margin than when it decreases performance.
This fact is hidden when simply looking at the global accuracy over the 42 DA labels, because the 10 best DAs for our model only correspond to 20.2\% of all annotations, whereas the 10 worst account for almost 40\% of all annotations.
\\

\noindent \textbf{Absolute differences}.
It is also interesting to note that the 10 DAs that our model best predicted are all very difficult DAs, for which the performance of the base model is very low in the first place: \textbf{31.7}, on average.
These DAs are also rare: they only correspond to 20.2\% of all annotations.
Our model raises the average accuracy on these labels to \textbf{37.08}.
On the other hand, the 10 DAs that are worst predicted by our model are more frequent DAs (40\% of all annotations), for which the performance of the base model is already quite high: \textbf{64.54}, on average.
And although our model is not as good as the base model on these DAs, it still reaches a decent average performance of \textbf{59.67}.
Therefore, our model provides a performance boost where it is most necessary (difficult and rare DAs), and wherever it fails, it still provides decent accuracy levels.\\

\noindent \textbf{10 best DAs for our model}.
Our model outperforms the base model by a very large margin of 20.0\% (20.0\%$\rightarrow$40.0\%) for Self-talk (\texttt{t1}, the speaker talks to him/herself).
It makes a lot of sense, as the accurate prediction of this DA obviously requires being aware of speaker-change.
Similar conclusions can be also drawn for Collaborative Completion (\texttt{\^{}2}, one speaker completes the other speaker's utterance), Repeat-phrase (\texttt{b\^{}m}, repeating parts of what the previous speaker said), Thanking (\texttt{ft}), Summarize/reformulate (\texttt{bf}, proposing a summarization or paraphrase of another speaker's talk/point), and Rhetorical-Questions (\texttt{qh}, questions asked to make a statement or asked to produce an effect with no answer expected).\\

\noindent \textbf{10 worst DAs for our model}.
On the other hand, speaker information does not seem to be crucial to predict the 10 DA labels most often missed by our model.
For instance, Conventional-openings (fp) are always found among the first utterances in a conversation, so there is only a small need for speaker-change awareness in that case.
E.g., in this situation with three utterances: (1) ``{\small \texttt{A: Hi, Wanet (fp)}}'', (2) ``{\small \texttt{A: How are you? (fp)}}'', and (3) ``{\small \texttt{B: I'm doing fine. (fp)}}'', utterances 2 and 3 are labeled with \texttt{fp}, regardless of speaker-change.
Likewise, the need for speaker-change awareness seems very little for the Quotation (\^{}q) and Other (ot) DAs.
In other words, among the DAs worst predicted by our model are DAs for which speaker information is not necessary to make an accurate prediction.
This makes sense, since the goal of our modified CRF layer is precisely to capture speaker information.

\section{DA label statistics}\label{app:da_label_statistics}

\begin{table*}[ht]
\small
\centering
\setlength{\tabcolsep}{2pt} 
\renewcommand{\arraystretch}{1.3} 
\scalebox{0.9}{
\begin{tabular}{|r|rr||r|rr|}
\hline
Dialogue Act (label) & count & frequency & Dialogue Act (label) & count & frequency \\ 
\hline
Statement-non-opinion (sd) & 73873 & 36.85\% & Collaborative Completion (\^{}2) & 709 & 0.35\% \\
Acknowledge/Backchannel (b) & 37727 & 18.82\% & Repeat-phrase (b\^{}m) & 677 & 0.34\% \\
Statement-opinion (sv) & 25810 & 12.88\% & Open-Question (qo) & 647 & 0.32\% \\
Abandoned/Uninterpretable (\%) & 15294 & 7.63\% & Rhetorical-Questions (qh) & 566 & 0.28\% \\
Agree/Accept (aa) & 10987 & 5.48\% & Hold before answer/agreement (\^{}h) & 546 & 0.27\% \\
Appreciation (ba) & 4702 & 2.35\% & Reject (ar) & 341 & 0.17\% \\
Yes-No-Question (qy) & 4679 & 2.33\% & Negative non-no answers (ng) & 296 & 0.15\% \\
Non-verbal (x) & 3565 & 1.78\% & Signal-non-understanding (br) & 295 & 0.15\% \\
Yes answers (ny) & 2995 & 1.49\% & Other answers (no) & 284 & 0.14\% \\
Conventional-closing (fc) & 2562 & 1.28\% & Conventional-opening (fp) & 225 & 0.11\% \\
Wh-Question (qw) & 1954 & 0.97\% & Or-Clause Question (qrr) & 208 & 0.10\% \\
No answers (nn) & 1363 & 0.68\% & Dispreferred answers (arp\_nd) & 207 & 0.10\% \\
Response Acknowledgement (bk) & 1299 & 0.65\% & 3rd-party-talk (t3) & 115 & 0.06\% \\
Hedge (h) & 1204 & 0.60\% & Offers, Options, Commits (oo\_co\_cc) & 109 & 0.05\% \\
Declarative Yes-No-Question (qy\^{}d) & 1203 & 0.60\% & Maybe/Accept-part (aap\_am) & 105 & 0.05\% \\
Backchannel in question form (bh) & 1036 & 0.52\% & Self-talk (t1) & 103 & 0.05\% \\
Quotation (\^{}q) & 948 & 0.47\% & Downplayer (bd) & 101 & 0.05\% \\
Summarize/reformulate (bf) & 928 & 0.46\% & Tag-Question (\^{}g) & 92 & 0.05\% \\
Other (ot) & 876 & 0.44\% & Declarative Wh-Question (qw\^{}d) & 80 & 0.04\% \\
Affirmative non-yes answers (na) & 841 & 0.42\% & Apology (fa) & 78 & 0.04\% \\
Action-directive (ad) & 740 & 0.37\% & Thanking (ft) & 74 & 0.04\% \\
\hline
\end{tabular}
}
\caption{Counts and frequencies of the 42 DA labels in the SwDA dataset. There  are 200444 utterances in total.}
\label{table:lables}
\end{table*}

\twocolumn
\bibliography{main}
\bibliographystyle{acl_natbib}
\end{document}